\documentclass[runningheads]{llncs}
\usepackage{times}
\usepackage{graphicx}
\usepackage{float}
\usepackage{makecell}
\usepackage{amsmath,amssymb} 
\usepackage{color}
\usepackage{xcolor,colortbl}
\definecolor{Gray}{gray}{0.9}

\usepackage{lipsum}

\newcommand\blfootnote[1]{%
  \begingroup
  \renewcommand\thefootnote{}\footnote{#1}%
  \addtocounter{footnote}{-1}%
  \endgroup
}
\DeclareMathOperator*{\argmin}{argmin}

\newcommand{\etal}{{\it et al. } }
\newcommand{\ie}{{\it i.e.,} }
\newcommand{\eg} {{\it e.g.,} }
\newcommand{\etc} {{\it etc.} }
\newcommand{\wrt} {w.r.t. }

\def\fig#1{Fig.~\ref{fig:#1}}

\begin{document}

\title{Adaptive Affinity Fields for Semantic Segmentation}
\titlerunning{Adaptive Affinity Fields}

\author{
Tsung-Wei Ke*
\and
Jyh-Jing Hwang*
\and
Ziwei Liu
\and
Stella X. Yu
}
\authorrunning{T.-W. Ke, J.-J. Hwang, Z. Liu, and S. X. Yu}

\institute{UC Berkeley / ICSI \\
\email{\{twke,jyh,zwliu,stellayu\}@berkeley.edu}\\
}

\maketitle

\blfootnote{* Equal contributors.  Jyh-Jing is a visiting student from the University of Pennsylvania.}

\begin{abstract}

Semantic segmentation has made much progress with increasingly powerful pixel-wise classifiers and incorporating structural priors via Conditional Random Fields (CRF) or Generative Adversarial Networks (GAN).
We propose a simpler alternative that learns to verify the spatial structure of segmentation {\it during training only}. 
Unlike existing approaches that enforce semantic labels on individual pixels and match labels between neighbouring pixels,  we propose the concept of {\it Adaptive Affinity Fields}  (AAF) to capture and {\it match the semantic relations} between neighbouring pixels in the label space.  
We use \textit{adversarial learning} to select the optimal affinity field size for each semantic category.  It is formulated as a $minimax$ problem, optimizing our segmentation neural network in a best worst-case learning scenario.
AAF is versatile for representing structures as a collection of pixel-centric relations, easier to train than GAN and more efficient than CRF without run-time inference.
Our extensive evaluations on PASCAL VOC 2012, Cityscapes, and GTA5 datasets demonstrate its above-par segmentation performance and robust generalization across domains.

\keywords{semantic segmentation; affinity field; adversarial learning}

\end{abstract}
\section{Introduction}
\label{sec:intro}

\begin{figure*}[tp]
    \centering
     \includegraphics[width=0.99\linewidth]{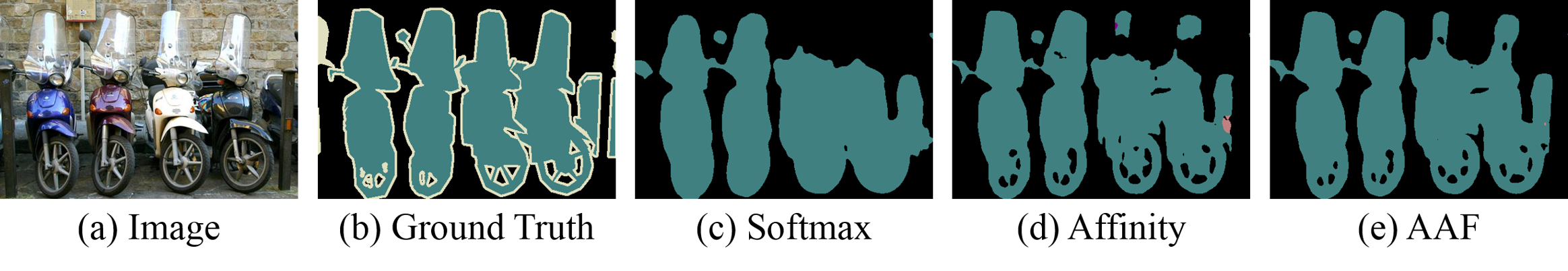}
     \caption{We propose new pairwise pixel loss functions that capture the spatial structure of segmentation.  Given an image ({\bf a}),  the task is to predict the ground-truth labeling ({\bf b}).  When a deep neural net is trained with conventional softmax cross-entropy loss on individual pixels, the predicted segmentation ({\bf c}) is often based on visual appearance and oblivious of the spatial structure of each semantic class.  Our work imposes an additional pairwise pixel label affinity loss ({\bf d}), matching the label relations among neighouring pixels between the prediction and the ground-truth.  We also learn the neighbourhood size for each semantic class, and our adaptive affinity fields result ({\bf e}) picks out both large bicycle shields and thin spokes of round wheels.
     }
     \label{fig:main}
\end{figure*}

Semantic segmentation of an image refers to the challenging task of assigning each pixel a categorical label, e.g., {\it motorcycle} or {\it person}.
Segmentation performance is often measured in a pixel-wise fashion, 
in terms of mean Intersection over Union (mIoU) across categories between 
the ground-truth (\fig{main}b) and the predicted label map (\fig{main}c).

Much progress has been made on segmentation with convolutional neural nets (CNN), mostly due to increasingly powerful pixel-wise classifiers, \eg VGG-16~\cite{simonyan2014very,long2015fully} and ResNet~\cite{he2016deep,wu2016high}, with the convolutional filters optimized by minimizing the average pixel-wise classification error over the image.

Even with big training data and with deeper and more complex network architectures, pixel-wise classification based approaches fundamentally lack the spatial discrimination power when foreground pixels and background pixels are close or mixed together: Segmentation is poor when the visual evidence for the foreground is weak, \eg glass motorcycle shields, or when the spatial structure is small, \eg thin radial spokes of all the wheels (\fig{main}c). 

There have been two main lines of efforts at incorporating structural reasoning into semantic segmentation:
 Conditional Random Field  (CRF) methods~\cite{krahenbuhl2011efficient,zheng2015conditional} 
and Generative Adversarial Network (GAN) methods \cite{goodfellow2014generative,luc2016semantic}.
\begin{enumerate}
\setlength{\parskip}{0pt}\setlength{\leftskip}{-0.5em}\setlength{\itemsep}{1mm}
\item CRF enforces label consistency between pixels measured by the similiarity in visual appearance (\eg raw pixel value).  An optimal labeling is solved via message passing algorithms~\cite{chen2015learning,liu2017deep}.  CRF is employed either as a post-processing step~\cite{krahenbuhl2011efficient,chen2016deeplab}, or as a plug-in module inside deep neural networks~\cite{zheng2015conditional,liu2015semantic}.   Aside from its time-consuming iterative inference routine, CRF is also sensitive to visual appearance changes.

\item GAN is a recent alternative for imposing structural regularity in the neural network output.  Specifically, the predicted label map is tested by a discriminator network on whether it resembles ground truth label maps in the training set.  GAN is notoriously hard to train, particularly prone to model instability and mode collapses \cite{radford2015unsupervised}.  
\end{enumerate}

We propose a simpler approach, by learning to verify the spatial structure of segmentation {\it during training only}.  Instead of enforcing semantic labels on individual pixels and matching labels between neighbouring pixels using CRF or GAN,  we propose the concept of {\it Adaptive Affinity Fields (AAF)} to capture and match the relations between neighbouring pixels in the label space.  How the semantic label of each pixel is related to those of neighboring pixels, \eg whether they are {\it same or different}, provides a distributed and pixel-centric description of semantic relations in the space and collectively they describe {\it Motorcycle wheels are round with thin radial spokes}.  We develop new affinity field matching loss functions to learn a CNN that automatically outputs a segmentation respectful of spatial structures and small details.

The pairwise pixel affinity idea has deep roots in perceptual organization, where local affinity fields have been used to characterize the intrinsic geometric structures in early vision~\cite{poggio1985early}, the grouping cues between pixels for image segmentation via spectral graph partitioning \cite{shi2000normalized}, and the object hypothesis for non-additive score verification in object recognition at the run time \cite{amir1998grouping}.

Technically, affinity fields at different neighbourhood sizes encode structural relations at different ranges.  Matching the affinity fields at a fixed size would not work well for all semantic categories, \eg thin structures are needed for {\it persons} seen at a distance whereas large structures are for {\it cows} seen close-up.

One straightforward solution is to search over a list of possible affinity field sizes, and pick the one that yields the minimal affinity matching loss.   However, such a practice would result in selecting trivial sizes which are readily satisfied.  For example, for large uniform semantic regions, the optimal affinity field size would be the smallest neighbourhood size of 1, and any pixel-wise classification would already get them right without any additional loss terms in the label space.

We propose \textit{adversarial learning} for size-adapted affinity field matching.  
Intuitively, we select the right size by pushing the affinity field matching with different sizes to the extreme: Minimizing the affinity loss should be hard enough to have a real impact on learning, yet it should still be easy enough for the  network to actually improve segmentation towards the ground-truth, \ie a best worst-case learning scenario.
Specifically, we formulate our AAF as a $minimax$ problem where we simultaneously {\it maximize} the affinity errors over multiple kernel sizes and {\it minimize} the overall matching loss.   Consequently, our adversarial network learns to assign a smaller affinity field size to {\it person} than to {\it cow}, as the person category contains finer structures than the cow category.

Our AAF has a few appealing properties over existing approaches (Table~\ref{tab:comparison}).

\begin{table}[h]
    \centering
    \resizebox{\textwidth}{!}{%
    \begin{tabular}{l|c|c|c|c}
    \Xhline{1pt}
    ~{\bf Method}~ & ~{\bf Structure Guidance}~ & ~{\bf Training}~ & ~{\bf Run-time Inference}~ & ~{\bf Performance}~ \\ \hline \hline
    ~CRF~\cite{krahenbuhl2011efficient}~ & input image & medium & yes & 76.53 \\ \hline
    ~GAN~\cite{goodfellow2014generative}~ & ground-truth labels & hard & no & 76.20 \\ \hline
    Our ~AAF~ & ~label affinity ~ & easy & no & {\bf 79.24} \\
    \Xhline{1pt}
    \end{tabular}}
    \vspace{0.5pt}
    \caption{Key differences between our method and other popular structure modeling approaches, namely CRF~\cite{krahenbuhl2011efficient} and GAN~\cite{goodfellow2014generative}. The performance (\% mIoU) is reported with PSPNet~\cite{zhao2016pyramid} architecture on the Cityscapes~\cite{Cordts2016Cityscapes} validation set. }
    \label{tab:comparison}
\end{table}

\begin{enumerate}
\setlength{\parskip}{0pt}\setlength{\leftskip}{-0.5em}\setlength{\itemsep}{1mm}

\item It provides a {\bf versatile representation} that encodes spatial structural information in distributed, pixel-centric relations.

\item It is {\bf easier to train} than GAN and {\bf more efficient} than CRF, as AAF only impacts network learning during training, requiring no extra parameters or inference processes during testing.

\item It is {\bf more generalizable to visual domain changes}, as AAF operates on the label relations not on the pixel values, capturing desired intrinsic geometric regularities despite of visual appearance variations.
\end{enumerate}

We demonstrate its effectiveness and efficiency with extensive evaluations on Cityscapes~\cite{Cordts2016Cityscapes} and PASCAL VOC 2012~\cite{everingham2010pascal} datasets, along with its remarkable generalization performance when our learned networks are applied to the GTA5 dataset ~\cite{Richter_2016_ECCV}.

\section{Related Works}
\label{sec:work}

Most methods treat semantic segmentation as a pixel-wise classification task, and those that model structural correlations provide a small gain at a large computational cost.

\noindent
\textbf{Semantic Segmentation.}
%
%
Since the introduction of fully convolutional networks for semantic segmentation ~\cite{long2015fully}, deeper~\cite{wu2016high,zhao2016pyramid,li2017not} and wider~\cite{noh2015learning,ronneberger2015u,yu2015multi} network architectures have been 
explored, drastically improving the performance on benchmarks such as PASCAL VOC~\cite{everingham2010pascal}.
For example, Wu \etal\cite{wu2016high} achieved higher segmentation accuracy by replacing backbone networks with more powerful ResNet~\cite{he2016deep}, whereas Yu \etal\cite{yu2015multi} tackled fine-detailed segmentation using atrous convolutions.
While the performance gain in terms of mIoU is impressive, these
pixel-wise classification based approaches fundamentally lack the spatial discrimination power when foreground and background pixels are close or mixed together, resulting in unnatural artifacts in Fig.~\ref{fig:main}c. 

\noindent
\textbf{Structure Modeling.}
%
%
%
%
Image segmentation has highly correlated outputs among the pixels.  Formulating it as an independent pixel labeling problem not only makes the pixel-level classification unnecessarily hard, but also leads to artifacts and spatially incoherent results.  Several ways to incorporate structure information into segmentation have been investigated~\cite{krahenbuhl2011efficient,chen2015learning,zheng2015conditional,liu2015semantic,lin2016efficient,bertasius2016convolutional,mostajabi2018regularizing}.
For example, Chen \etal\cite{chen2016deeplab} utilized denseCRF~\cite{krahenbuhl2011efficient} as post-processing to refine the final segmentation results.
Zheng \etal\cite{zheng2015conditional} and Liu \etal\cite{liu2015semantic} further made the CRF module differentiable within the deep neural network.
Pairwise low-level image cues, such as grouping affinity~\cite{maire2016affinity,liu2017learning} and contour cues~\cite{bertasius2016semantic,chen2016semantic}, have also been used to encode structures.
However, these methods are sensitive to visual appearance changes, or require expensive iterative inference procedures.

Our work provides another perspective to structure modeling by matching the relations between neighbouring pixels in the label space.   Our segmentation network learns to verify the spatial structure of segmentation only during training; once it is trained, it is ready for deployment without run-time inference.

\section{Our Approach: Adaptive Affinity Fields}
\label{sec:method}

\begin{figure}[t]
    \centering
    \includegraphics[width=1\textwidth]{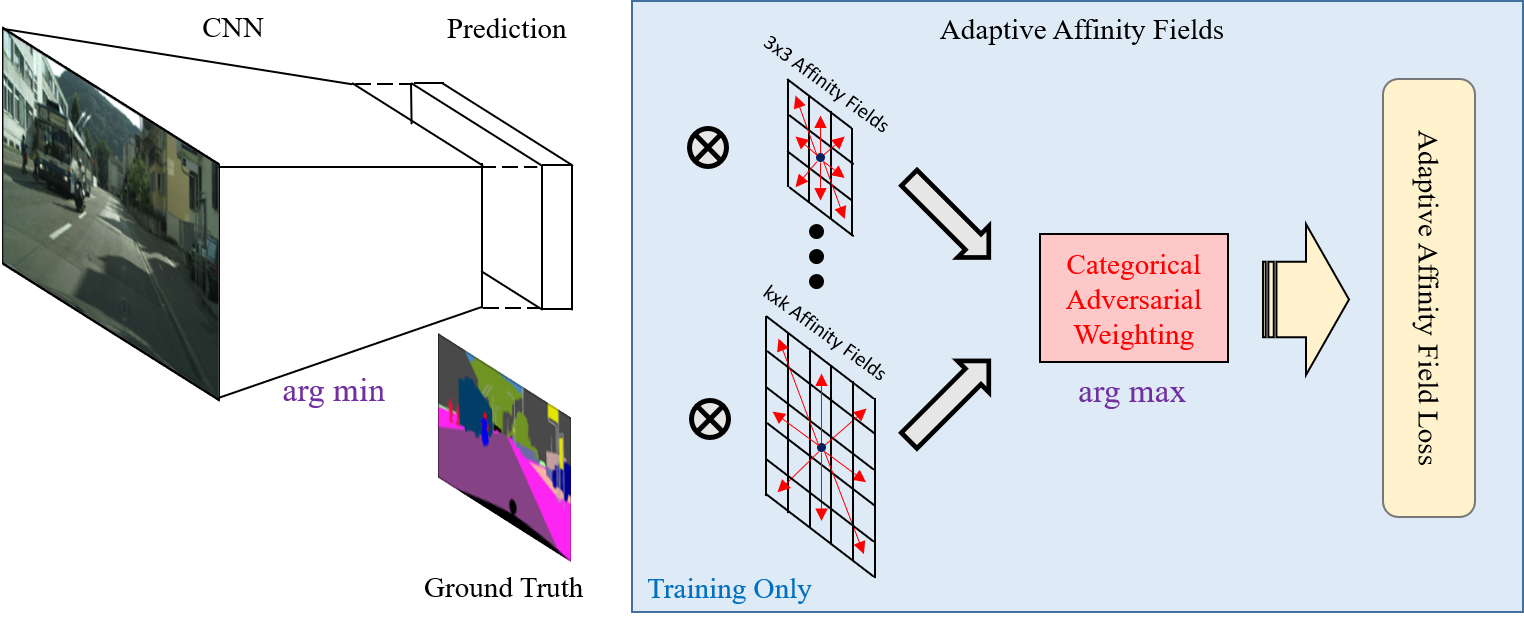}
    \caption{Method overview: Learning semantic segmentation with adaptive affinity fields.  The adaptive affinity fields consist of two parts: the affinity field loss with multiple kernel sizes and corresponding categorical adversarial weightings.  Note that the adaptive affinity fields are only introduced during training and there is no extra computation during inference.}
    \label{fig:architecture}
\end{figure}

We first briefly revisit the classic pixel-wise cross-entropy loss commonly used in semantic segmentation. The drawbacks of pixel-wise supervision lead to our concept of region-wise supervision. We then describe our region-wise supervision through affinity fields, and introduce an adversarial process that learns an adaptive affinity kernel size for each category. We summarize the overall AAF architecture in Fig.~\ref{fig:architecture}.

\subsection{From Pixel-wise Supervision to Region-wise Supervision}

Pixel-wise cross-entropy loss is most often used in CNNs for semantic segmentation~\cite{long2015fully,chen2016deeplab}.  It penalizes pixel-wise predictions independently and is known as a form of {\it unary supervision}.   It implicitly assumes that the relationships between pixels can be learned as the effective receptive field increases with deeper layers.  Given predicted categorical probability $\hat{y}_i(l)$ at pixel $i$ \wrt its ground truth categorical label $l$, the total loss is the average of cross-entropy loss at pixel $i$:
\begin{equation}
   \mathcal{L}_\text{unary}^i =  \mathcal{L}_\text{cross-entropy}^i = -\log \hat{y}_i(l).
\end{equation}
Such a unary loss does not take the semantic label correlation and scene structure into account.
The objects in different categories interact with each other in a certain pattern. For example, cars are usually on the road while pedestrians on the sidewalk; buildings are surrounded by the sky but never on top of it. Also, some shapes of a certain category occur more frequently, such as rectangles in trains, circles in bikes, and straight vertical lines in poles. This kind of inter-class and inner-class pixel relationships are informative and can be integrated into learning as structure reasoning. We are thus inspired to propose an additional region-wise loss to impose penalties on inconsistent unary predictions and encourage the network to learn such intrinsic pixel relationships. 

Region-wise supervision extends its pixel-wise counterpart from independent pixels to neighborhoods of pixels, \ie, the region-wise loss considers a patch of predictions and ground truth jointly. Such region-wise supervision $\mathcal{L}_\text{region}$ involves designing a specific loss function for a patch of predictions $\mathcal{N}(\hat{y_i})$ and corresponding patch of ground truth $\mathcal{N}({y_i})$ centered at pixel $i$, where $\mathcal{N}(\cdot)$ denotes the neighborhood.

The overall objective is hence to minimize the combination of unary and region losses, balanced by a constant $\lambda$:
\begin{equation}
    S^* = \argmin_S \mathcal{L} = \argmin_S \frac{1}{n}\sum_i\Big( \mathcal{L}_\text{unary}^i(\hat{y_i}, y_i) + \lambda \mathcal{L}_\text{region}^i\big(\mathcal{N}(\hat{y_i}), \mathcal{N}(y_i)\big) \Big),
\end{equation}
where $n$ is the total number of pixels. We omit index $i$ and averaging notations for simplicity in the rest of the paper.

The benefits of the addition of region-wise supervision have been explored in previous works.  For example, Luc \etal\cite{luc2016semantic} exploited GAN~\cite{goodfellow2014generative} as structural priors, and Mostajabi \etal\cite{mostajabi2018regularizing} pre-trained an additional auto-encoder to inject structure priors into training the segmentation network.  However, their approaches require much hyper-parameter tuning and are prone to overfitting, resulting in very small gains over strong baseline models.  Please see Table~\ref{tab:comparison} for a comparison. 

\subsection{Affinity Field Loss Function}
\label{sec:affinity}

Our affinity field loss function overcome these drawbacks and is a flexible region-wise supervision approach that is also easy to optimize.

The use of pairwise pixel affinity has a long history in image segmentation ~\cite{shi2000normalized,stella2003multiclass}.   The grouping relationships between neighbouring pixels are derived from the image and represented by a graph, where a node denotes a pixel and a weighted edge between two nodes captures the similarity between two pixels.  Image segmentation then becomes a graph partitioning problem, where all the nodes are divided into disjoint sets, with maximal weighted edges within the sets and minimal weighted edges between the sets.

\begin{figure}[t]
    \centering
    \includegraphics[width=1.0\textwidth]{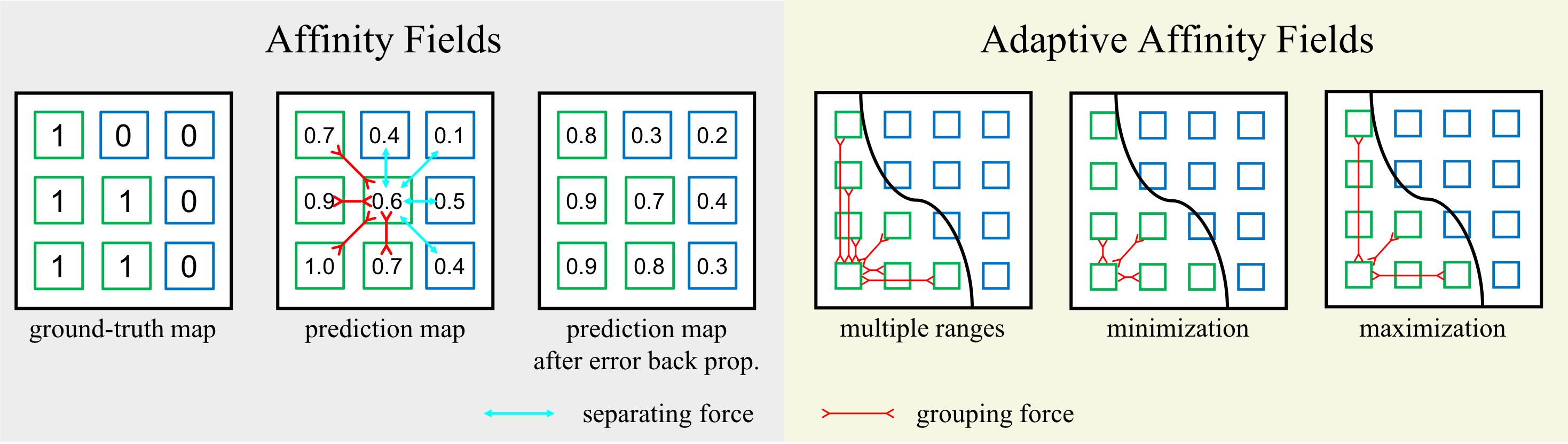}
    \caption{{\bf Left:} Our affinity field loss separates predicted probabilities across the boundary and unifies them within the segment. {\bf Right:} The affinity fields can be defined over multiple ranges.  Minimizing the affinity loss over different ranges results in trivial solutions which are readily satisfied.  
Our size-adaptive affinity field loss is achieved with adversarial learning: Maximizing the affinity loss over different kernel sizes selects the most critical range for imposing pairwise relationships in the label space, and our goal is to minimize this maximal loss -- i.e., use the best worst case scenario for most effective training.
}
    \label{fig:details}
\end{figure}

We define pairwise pixel affinity based not on the image, but on ground-truth label map.
There are two types of label relationships between a pair of pixels: whether their labels are the same or different.  
If pixel $i$ and its neighbor $j$ have the same categorical label, we impose a grouping force which encourages network predictions at $i$ and $j$ to be similar.  Otherwise, we impose a separating force which pushes apart their label predictions. These two forces are illustrated in Fig.~\ref{fig:details} left.

Specifically, we define a pairwise affinity loss based on KL divergence between binary classification probabilities, consistent with the cross-entropy loss for the unary label prediction term.  For pixel $i$ and its neighbour $j$, depending on whether two pixels belong to the same category $c$ in the ground-truth label map $y$, we define a non-boundary term $\mathcal{L}_{\text{affinity}}^{i\bar{b}c}$  for the grouping force and an boundary term $\mathcal{L}_{\text{affinity}}^{ibc}$ for the separating force in the prediction map $\hat{y}$:
\begin{equation}
    \mathcal{L}_{\text{affinity}}^{ic} = 
    \begin{cases}
    \mathcal{L}_{\text{affinity}}^{i\bar{b}c} = D_{KL}(\hat{y}_j(c)|| \hat{y}_i(c)) & \text{if } y_i(c) = y_j(c) \\
    \mathcal{L}_{\text{affinity}}^{ibc} = \max\{0, m - D_{KL}(\hat{y}_j(c) || \hat{y}_i(c))\} & \text{otherwise}
    \end{cases}
\end{equation}
$D_{KL}(\cdot)$ is the Kullback-Leibler divergence between two Bernoulli distributions $P$ and $Q$ with parameters $p$ and $q$ respectively: $D_{KL}(P||Q)= p\log \frac{p}{q}+\bar{p}\log \frac{\bar{p}}{\bar{q}}$ for the binary distribution $[p,1-p]$ and $[q,1-q]$, where $p,q \in [0,1]$.  For simplicity, we abbreviate the notation as $D_{KL}(p||q)$.  $\hat{y}_j(c)$ denotes the prediction probability of $j$ in class $c$.  The overall loss is the average of $\mathcal{L}_{\text{affinity}}^{ic}$ over all categories and pixels.

\noindent
{\bf Discussion 1.}  Our affinity loss encourages similar network predictions on two pixels of the same ground-truth label, regardless of what their actual labels are.
The collection of such pairwise bonds inside a segment ensure that all the pixels achieve the same label.  On the other hand, our affinity loss pushes network predictions apart on two pixels of different ground-truth labels, again regardless of what their actual labels are.  The collection of such pairwise repulsion help create clear segmentation boundaries.  

\noindent
{\bf Discussion 2.}  Our affinity loss may appear similar to CRF~\cite{krahenbuhl2011efficient} on the pairwise grouping or separating forces between pixels.  However, a crucial difference is that CRF models require iterative inference to find a solution, whereas our affinity loss only impacts the network training with pairwise supervision.  A similar perspective is metric learning with contrastive loss~\cite{chopra2005learning}, commonly used in face identification tasks.  Our affinity loss works better for segmentation tasks, because it penalizes the network predictions directly, and our pairwise supervision is {\it in addition to} and {\it consistent with} the conventional unary supervision. 

\subsection{Adaptive Kernel Sizes from Adversarial Learning}
\label{sec:aaf}

Region-wise supervision often requires a preset kernel size for CNNs, where pairwise pixel relationships are measured in the same fashion across all pixel locations.  However, we cannot expect one kernel size fits all categories, since the ideal kernel size for each category varies with the average object size and the object shape complexity.  

We propose a size-adaptive affinity field loss function, optimizing the weights over a set of affinity field sizes for each category in the loop:
\begin{equation}
    \mathcal{L}_\text{multiscale} = \sum_c \sum_k w_{ck} \mathcal{L}_\text{region}^{ck} \text{ \ \ s.t. } \sum_k w_{ck} = 1 \text{\ and \ } w_{ck} \ge 0
\end{equation}
where $\mathcal{L}_\text{region}^{ck}$ is a region loss defined in Eqn. (2), yet operating on a specific class channel $c$ with kernel size $k\times k$ with a corresponding weighting $w_{ck}$.

If we just minimize the affinity loss with size weighting $w$ included, $w$ would likely fall into a trivial solution. As illustrated in Fig~\ref{fig:details} right, the affinity loss would be minimum if the smallest kernels are highly weighted for non-boundary terms and the largest kernels for boundary terms, since nearby pixels are more likely to belong to the same object and far-away pixels to different objects.  Unary predictions based on the image would naturally have such statistics, nullifying any potential effect from our pairwise affinity supervision.

To optimize the size weighting without trivializing the affinity loss, we need to push the selection of kernel sizes to the extreme.  Intuitively, we need to enforce pixels in the same segment to have the same label prediction as far as possible, and likewise to enforce pixels in different segments to have different predictions as close as possible.  We use the best worst case scenario for most effective training.

We formulate the adaptive kernel size selection process as optimizing a two-player minimax game: While the segmenter should always attempt to minimize the total loss, the weighting for different kernel sizes in the loss should attempt to maximize the total loss in order to capture the most critical neighbourhood sizes.  Formally, we have: 
\begin{equation}
    S^* = \argmin_S \max_w \mathcal{L}_\text{unary} + \mathcal{L}_\text{multiscale}.
\end{equation}
For our size-adaptive affinity field learning, we separate the non-boundary term $\mathcal{L}_\text{affinity}^{\bar{b}ck}$ and boundary term $\mathcal{L}_\text{affinity}^{bck}$ in Eqn (3) since their ideal kernel sizes would be different. 
Our adaptive affinity field (AAF) loss becomes:
\begin{align}
    S^* &= \argmin_S \max_w \mathcal{L}_\text{unary} + \mathcal{L}_\text{AAF},\\
    \mathcal{L}_\text{AAF} &= \sum_c \sum_k (w_{\bar{b}ck} \mathcal{L}_\text{affinity}^{\bar{b}ck} + w_{bck} \mathcal{L}_\text{affinity}^{bck}),\\ 
    \nonumber 
\text{s.t. } \sum_k w_{\bar{b}ck} &= \sum_k w_{bck}= 1 \text{\ and \ } w_{\bar{b}ck},w_{bck} \ge 0.
\end{align}

\section{Experimental Setup}
\label{sec:exp_setup}

\subsection{Datasets}
\label{subsec:datasets}
We compare our proposed affinity fields and AAF with other competing methods on the PASCAL VOC 2012~\cite{everingham2010pascal} and Cityscapes~\cite{Cordts2016Cityscapes} datasets.

\noindent
\textbf{PASCAL VOC 2012.}
\label{sec:voc}
PASCAL VOC 2012~\cite{everingham2010pascal} segmentation dataset contains 20 object categories and one background class. Following the procedure of~\cite{long2015fully}, ~\cite{zhao2016pyramid}, ~\cite{chen2016deeplab}, we use augmented data with the annotations of~\cite{hariharan2011semantic}, resulting in 10,582, 1,449, and 1,456 images for training, validation and testing. 

\noindent
\textbf{Cityscapes.}
\label{sec:cityscapes}
Cityscapes~\cite{Cordts2016Cityscapes} is a dataset for semantic urban street scene understanding.  5000 high quality
pixel-level finely annotated images are divided into  training, validation, and testing sets with 2975, 500, and 1525 images, respectively. It defines 19 categories containing flat, human, vehicle, construction, object, nature, \etc

\subsection{Evaluation Metrics}
\label{subsec:metrics}
All existing semantic segmentation works adopt \textbf{pixel-wise mIoU}~\cite{long2015fully} as their metric.  To fully examine the effectiveness of our AAF on fine structures in particular, we also evaluate all the models using \textbf{instance-wise mIoU} and \textbf{boundary detection metrics}.

\noindent
\textbf{Instance-wise mIoU.}
Since the pixel-wise mIoU metric is often biased toward large objects, we introduce the instance-wise mIoU to alleviate the bias, which allow us to evaluate fairly the performance on smaller objects. The per category instance-wise mIoU is formulated as $\hat{U}_c = \frac{\sum_x n_{c,x} \times U_{c,x}}{\sum_x n_{c,x}},$
where $n_{c,x}$ and $U_{c,x}$ are the number of instances and IoU of class $c$ in image $x$, respectively. 

\noindent
\textbf{Boundary detection metrics.}
We compute semantic boundaries using the semantic predictions and benchmark the results using the standard benchmark for contour detection proposed by~\cite{amfm_pami2011}, which summarizes the results by precision, recall, and f-measure. 

\subsection{Methods of Comparison}
\label{subsec:baseline_methods}
We briefly describe other popular methods that are used for comparison in our experiments, namely, GAN's adversarial learning~\cite{goodfellow2014generative}, contrastive loss~\cite{chopra2005learning}, and CRF~\cite{krahenbuhl2011efficient}. \\

\noindent
\textbf{GAN's Adversarial Learning.} 
We investigate a popular framework, the Generative Adversarial Networks (GAN)~\cite{goodfellow2014generative}. The discriminator $D$ in GAN works as injecting priors for region structures. The adversarial loss is formulated as
\begin{equation}
    \mathcal{L}_{\text{adversarial}}^i = \log D(\mathcal{N}(y_i)) + \log(1-D(\mathcal{N}(\hat{y_i}))).
\end{equation}
We simultaneously train the segmentation network $S$ to minimize $\log (1-D(\mathcal{N}(\hat{y_i})))$ and the discriminator to maximize $\mathcal{L}_{\text{adversarial}}$. \\

\noindent
\textbf{Pixel Embedding.} 
We study the region-wise supervision over feature map, which is implemented by imposing the contrastive loss~\cite{chopra2005learning} on the last convolutional layer before the softmax layer. The contrastive loss is formulated as

\begin{equation}
    \mathcal{L}_{\text{contrast}}^{i} = 
    \begin{cases}
    \mathcal{L}_{\text{contrast}}^{i\bar{e}} = \|f_j-f_i\|^2_2 & \text{if } y_i(c) = y_j(c) \\
    \mathcal{L}_{\text{contrast}}^{ie} = \max\{0, m - \|f_j-f_i\|^2_2\} & \text{otherwise,}
    \end{cases}
\end{equation}
where $f_i$ denotes $L_2$-normalized feature vector at pixel $i$, and $m$ is set to $0.2$. \\

\noindent 
\textbf{CRF-based Processing.} 
We follow ~\cite{chen2016deeplab}'s implementation by post-processing the prediction with dense-CRF~\cite{krahenbuhl2011efficient}. We set $bi\_w$ to $1$, $bi\_xy\_std$ to $40$, $bi\_rgb\_std$ to $3$, $pos\_w$ to $1$, and $pos\_xy\_std$ to $1$ for all experiments. It is worth mentioning that CRF takes additional $40$ seconds to generate the final results on Cityscapes, while our proposed methods introduce no inference overhead.

\subsection{Implementation Details}
\label{subsec:imp_details}

Our implementation follows the ones of base architectures, which are PSPNet~\cite{zhao2016pyramid} in most cases or FCN~\cite{long2015fully}. We use the poly learning rate policy where the current learning rate equals the base one multiplied by  $(1-\frac{\text{iter}}{\text{max\_iter}})^{0.9}$. We set the base learning rate as $0.001$. The training iterations for all experiments is $30K$ on VOC dataset and $90K$ on Cityscapes dataset while the performance can be further improved by increasing the iteration number. Momentum and weight decay are set to $0.9$ and $0.0005$, respectively.  For data augmentation, we adopt random mirroring and random resizing between 0.5 and 2 for all
datasets. We do not upscale the logits (prediction map) back to the input image resolution, instead, we follow \cite{chen2016deeplab}'s setting by downsampling the ground-truth labels for training ($output\_stride=8$).

PSPNet~\cite{zhao2016pyramid} shows that larger ``cropsize'' and ``batchsize'' can yield better performance. In their implementation, ``cropsize'' can be up to $720\times 720$ and ``batchsize'' to $16$ using $16$ GPUs. To speed up the experiments for validation on VOC, we downsize ``cropsize'' to $336\times 336$ and ``batchsize'' to $8$ so that a single GTX Titan X GPU is sufficient for training.  We set ``cropsize'' to $480\times 480$ during inference. For testing on PASCAL VOC 2012 and all experiments on Cityscapes dataset, we use $4$-GPUs to train the network. On VOC dataset, we set the ``batchsize'' to 16 and set ``cropsize'' to $480 \times 480$. On Cityscaeps, we set the ``batchsize'' to 8 and ``cropsize'' to $720 \times 720$. For inference, we boost the performance by averaging scores from left-right flipped and multi-scale inputs ($scales = \{0.5,0.75,1,1.25,1.5,1.75\}$). 

For affinity fields and AAF, $\lambda$ is set to $1.0$ and margin $m$ is set to $3.0$. We use ResNet101~\cite{he2016deep} as the backbone network and initialize the models with weights pre-trained on ImageNet~\cite{ILSVRC15}.

\begin{table*}[b]
  \centering
  \resizebox{\textwidth}{!}{%
  \begin{tabular}{l|c c c c c c c c c c c c c c c c c c c c|c}
    \Xhline{1pt}
    Method & aero & bike & bird & boat & bottle & bus & car & cat & chair & cow & table & dog & horse & mbike & person & plant & sheep & sofa & train & tv & mIoU \\
    \hline \hline
    \rowcolor{Gray}
    FCN & 86.95 & 59.25 & 85.18 & 70.33 & 73.92 & 78.86 & 82.30 & 85.64 & 33.57 & 69.34 & 27.41 & 78.04 & 71.45 & 70.45 & 85.54 & 57.42 & 71.55 & 32.48 & 74.91 & 59.10 & 68.91 \\
    PSPNet & 92.56 & 66.70 & 91.10 & 76.52 & 80.88 & 94.43 & 88.49 & 93.14 & 38.87 & 89.33 & 62.77 & 86.44 & 89.72 & 88.36 & 87.48 & 56.95 & 91.77 & 46.23 & 88.59 & 77.14 &     80.12 \\
    \hline \hline
    \rowcolor{Gray}
    Affinity & 88.66 & 59.25 & 87.85 & 72.19 & 76.36 & 80.65 & 80.74 & 87.82 & 35.38 & 73.45 & 30.17 & 79.84 & 68.15 & 73.52 & 87.96 & 53.95 & 75.46 & 37.15 & 76.62 & 73.42 & 71.07 \\
    
    \rowcolor{Gray}
    AAF & 88.15 & 67.83 & 87.06 & 72.05 & 76.45 & 85.43 & 80.58 & 88.33 & 35.47 & 72.76 & 31.55 & 79.68 & 67.01 & 77.96 & 88.20 & 50.31 & 73.16 & 42.71 & 78.14 & 73.87 & 71.95 \\
    
    GAN & 92.36 & 65.94 & 91.80 & 76.35 & 77.70 & 95.39 & 89.21 & 93.30 & 43.35 & 89.25 & 61.81 & 86.93 & 91.28 & 87.43 & 87.21 & 68.15 & 90.64 & 49.64 & 88.79 & 73.83 & 80.74 \\  

    Emb. & 91.28 & 69.50 & 92.62 & 77.60 & 78.74 & 95.03 & 89.57 & 93.67 & 43.21 & 88.76 & 62.47 & 86.68 & 91.28 & 88.47 & 87.44 & 69.21 & 91.53 & 52.17 & 89.30 & 74.60 & 81.36 \\
    
    Affinity & 91.52 & {\bf 74.74} & 92.09 & 78.17 & 80.73 & 95.70 & 89.52 & 92.83 & 43.29 & 89.21 & 60.33 & 87.50 & 90.96 & 88.77 & 88.88 & \textbf{71.00} & 88.54 & 50.61 & 89.64 & 78.22 & 81.80 \\

    AAF & 92.97 & 73.68 & 92.49 & 80.51 & 79.73 & 96.15 & 90.92 & 93.42 & \textbf{45.11} & 89.00 & 62.87 & 87.97 & 91.32 & 90.28 & \textbf{89.30} & 69.05 & 88.92 & \textbf{52.81} & 89.05 & 78.91 & \textbf{82.39} \\
    \Xhline{1pt}
    \end{tabular}}
    \vspace{0.5pt}
    \caption{Per-class results on Pascal VOC 2012 validation set. Gray colored background denotes using FCN as the base architecture.}
    \label{tab:voc}
\end{table*}

\begin{table*}[t]
  \centering
  \resizebox{\textwidth}{!}{%
  \begin{tabular}{l|c c c c c c c c c c c c c c c c c c c|c}
    \Xhline{1pt}
    Method & road & swalk & build. & wall & fence & pole & tlight & tsign & veg. & terrain & sky & person & rider & car & truck & bus & train & mbike & bike & mIoU \\
    \hline \hline
    \rowcolor{Gray}
    FCN & 97.31 & 79.28 & 89.52 & 38.08 & 48.63 & 49.70 & 59.37 & 69.94 & 90.86 & 56.58 & 92.38 & 75.91 & 46.24 & 92.26 & 50.41 & 64.51 & 39.73 & 54.91 & 73.07 & 66.77 \\

    PSPNet & 97.96 & 83.89 & 92.22 & 57.24 & 59.31 & 58.89 & 68.39 & 77.07 & 92.18 & 63.71 & 94.42 & 81.80 & 63.11 & 94.85 & 73.54 & 84.82 & 67.42 & 69.34 & 77.42 & 76.72 \\
    \hline \hline
    \rowcolor{Gray}
    Affinity & 97.52 & 80.90 & 90.42 & 40.45 & 49.81 & 55.97 & 63.92 & 73.37 & 91.49 & 59.01 & 93.30 & 78.17 & 52.16 & 92.85 & 52.53 & 65.78 & 39.28 & 52.88 & 74.53 & 68.65 \\
    
    \rowcolor{Gray}
    AAF & 97.58 & 81.19 & 90.50 & 42.30 & 50.34 & 57.47 & 65.39 & 74.83 & 91.54 & 59.25 & 93.11 & 78.65 & 52.98 & 93.15 & 53.10 & 67.58 & 38.40 & 51.57 & 74.80 & 69.14 \\

    CRF & 97.96 & 83.82 & 92.14 & 57.16 & 59.28 & 57.48 & 67.71 & 76.61 & 92.09 & 63.67 & 94.35 & 81.62 & 62.98 & 94.81 & 73.59 & 84.81 & 67.49 & 69.22 & 77.28 & 76.53 \\
    
    GAN & 97.95 & 83.59 & 92.01 & 56.92 & 60.17 & 58.63 & 68.37 & 77.36 & 92.28 & 62.70 & 94.42 & 81.59 & 62.27 & 94.94 & 78.09 & 82.79 & 56.75 & 69.19 & 77.78 & 76.20 \\
    
    Affinity & 98.08 & 85.58 & 92.60 & 58.33 & 61.45 & \textbf{66.80} & \textbf{74.19} & 81.29 & 92.90 & 65.34 & 94.87 & 84.00 & 65.84 & 95.50 & 76.84 & 85.80 & 64.19 & 72.32 & 79.83 & 78.72 \\

    AAF & 98.18 & 85.35 & 92.86 & 58.87 & 61.48 & 66.64 & 74.00 & 80.98 & 92.95 & 65.31 & 94.91 & \textbf{84.27} & \textbf{66.98} & 95.51 & 79.39 & 87.06 & 67.80 & 72.91 & 80.19 & \textbf{79.24} \\
    \Xhline{1pt}
    \end{tabular}}
    \vspace{0.5pt}
    \caption{Per-class results on Cityscapes validation set. Gray colored background denotes using FCN as the base architecture.}
    \label{tab:cityscapes}
\end{table*}

\section{Experimental Results}
\label{sec:exp_results}
We benchmark our proposed methods on two datasets, PASCAL VOC 2012~\cite{everingham2010pascal} and Cityscapes~\cite{Cordts2016Cityscapes}. All methods are evaluated by three metrics: mIoU, instance-wise mIoU and boundary detection recall. We include some visual examples to demonstrate the effectiveness of our proposed methods in Fig.~\ref{fig:results}.

\subsection{Pixel-level Evaluation}
\label{sec:pix_eval}

\noindent
\textbf{Validation Results.}
For training on PASCAL VOC 2012~\cite{everingham2010pascal}, we first train on $train\_aug$ for 30K iterations and then fine-tune on $train$ for another 30K iterations with base learning rate as $0.0001$. For Cityscapes~\cite{Cordts2016Cityscapes}, we only train on finely annotated images for 90K iterations. We summarize the mIoU results on validation set in Table~\ref{tab:voc} and Table~\ref{tab:cityscapes}, respectively.

With FCN~\cite{long2015fully} as base architecture, the affinity field loss and AAF improve the performance by $2.16\%$ and $3.04\%$ on VOC and by $1.88\%$ and $2.37\%$ on Cityscapes. With PSPNet~\cite{zhao2016pyramid} as the base architecture, the results also improves consistently: GAN loss, embedding contrastive loss, affinity field loss and AAF improve the mean IoU by $0.62\%$, $1.24\%$, $1.68\%$ and $2.27\%$ on VOC; affinity field loss and AAF improve by $2.00\%$ and $2.52\%$ on Cityscapes. It is worth noting that large improvements over PSPNet on VOC are mostly in categories with fine structures, such as ``bike'', ``chair'', ``person'', and ``plant''.

\noindent
\textbf{Testing Results.}
On PASCAL VOC 2012, the training procedure for PSPNet and AAF is the same as follows: We first train the networks on $train\_aug$ and then fine-tune on $train\_val$. We report the testing results on VOC 2012 and Cityscapes in Table~\ref{tab:voc_test} and Table~\ref{tab:cityscapes_test}, respectively. Our re-trained PSPnet does not reach the same performance as originally reported in the paper because we do not bootstrap the performance by fine-tuning on hard examples (like ``bike'' images), as pointed out in \cite{chen2017rethinking}. We demonstrate that our proposed AAF achieve $82.17\%$ and $79.07\%$ mIoU, which is better than the PSPNet by $1.54\%$ and $2.77\%$ and competitive to the state-of-the-art performance.\\

\begin{table*}[t]
  \centering
  \resizebox{\textwidth}{!}{%
  \begin{tabular}{l|c c c c c c c c c c c c c c c c c c c c|c}
    \Xhline{1pt}
    Method & aero & bike & bird & boat & bottle & bus & car & cat & chair & cow & table & dog & horse & mbike & person & plant & sheep & sofa & train & tv & mIoU \\
    \hline \hline

    PSPNet & 94.01 & 68.08 & 88.80 & 64.87 & 75.87 & 95.60 & 89.59 & 93.15 & 37.96 & 88.20 & 72.58 & 89.96 & 93.30 & 87.52 & 86.65 & 61.90 & 87.05 & 60.81 & 87.13 & 74.65 &     80.63 \\

    
    AAF & 91.25 & \textbf{72.90} & 90.69 & 68.22 & 77.73 & 95.55 & 90.70 & 94.66 & \textbf{40.90} & 89.53 & 72.63 & 91.64 & 94.07 & 88.33 & \textbf{88.84} & \textbf{67.26} & 92.88 & 62.62 & 85.22 & 74.02 & \textbf{82.17} \\
    \Xhline{1pt}
    \end{tabular}}
    \vspace{0.5pt}
    \caption{Per-class results on Pascal VOC 2012 testing set.}
    \label{tab:voc_test}
\end{table*}

\begin{table*}[b!]
  \centering
  \resizebox{\textwidth}{!}{%
  \begin{tabular}{l|c c c c c c c c c c c c c c c c c c c|c}
    \Xhline{1pt}
    Method & road & swalk & build. & wall & fence & pole & tlight & tsign & veg. & terrain & sky & person & rider & car & truck & bus & train & mbike & bike & mIoU \\
    \hline \hline
    PSPNet & 98.33 & 84.21 & 92.14 & 49.67 & 55.81 & 57.62 & 69.01 & 74.17 & 92.70 & 70.86 & 95.08 & 84.21 & 66.58 & 95.28 & 73.52 & 80.59 & 70.54 & 65.54 & 73.73 & 76.30 \\
    \hline
    AAF & 98.53 & 85.56 & 93.04 & 53.81 & 58.96 & \textbf{65.93} & 75.02 & 78.42 & 93.68 & 72.44 & 95.58 & 86.43 & 70.51 & 95.88 & 73.91 & 82.68 & 76.86 & 68.69 & 76.40 & \textbf{79.07} \\
    \Xhline{1pt}
    \end{tabular}}
    \vspace{0.5pt}
    \caption{Per-class results on Cityscapes test set.}
    \label{tab:cityscapes_test}
\end{table*}

\begin{table*}[b!]
  \centering
  \resizebox{\textwidth}{!}{%
  \begin{tabular}{l|c c c c c c c c c c c c c c c c c c c c|c}
    \Xhline{1pt}
    Method & aero & bike & bird & boat & bottle & bus & car & cat & chair & cow & table & dog & horse & mbike & person & plant & sheep & sofa & train & tv & mIoU \\
    \hline \hline
    PSPNet & 87.54 & 53.08 & 83.53 & 76.95 & 45.13 & 87.68 & 68.77 & 89.01 & 39.26 & 88.78 & 51.49 & 88.88 & 84.41 & 85.95 & 77.60 & 48.68 & 86.25 & 54.18 & 88.25 & 66.11 &     73.60 \\
    \hline \hline

    Affinity & 89.42 & 61.72 & 84.64 & 79.86 & 57.57 & 88.81 & 71.74 & 88.91 & 44.78 & 89.55 & 52.55 & 91.22 & 86.12 & 87.40 & 81.10 & 58.33 & 85.15 & 60.61 & 88.47 & 68.86 & 76.73 \\

    AAF & 89.76 & \textbf{61.74} & 84.40 & \textbf{81.87} & \textbf{58.04} & 89.03 & \textbf{73.68} & 90.46 & \textbf{46.67} & 89.65 & \textbf{55.63} & 91.33 & 85.85 & 88.36 & \textbf{81.93} & \textbf{59.84} & 84.52 & \textbf{62.67} & 89.35 & 68.80 & \textbf{77.54} \\
    \Xhline{1pt}
    \end{tabular}}
    \vspace{0.5pt}
    \caption{Per-class instance-wise IoU results on Pascal VOC 2012 validation set.}
    \label{tab:voc_inst}
\end{table*}

\begin{table*}[t]
  \centering
  \resizebox{\textwidth}{!}{%
  \begin{tabular}{l|c c c c c c c c c c c c c c c c c c c|c}
    \Xhline{1pt}
    Method & road & swalk & build. & wall & fence & pole & tlight & tsign & veg. & terrain & sky & person & rider & car & truck & bus & train & mbike & bike & mIoU \\
    \hline \hline
    PSPNet & 97.64 & 78.23 & 88.36 & 34.48 & 42.00 & 51.68 & 50.71 & 68.29 & 89.65 & 40.14 & 86.63 & 78.35 & 75.91 & 92.09 & 87.28 & 90.85 & 62.74 & 85.33 & 73.02 & 72.28\\
    \hline \hline

    Affinity & 97.73 & 80.51 & 89.32 & 38.21 & 45.89 & \textbf{61.31} & \textbf{59.75} & 73.41 & 90.62 & 43.22 & 88.20 & 81.18 & \textbf{80.29} & 93.24 & 89.60 & 94.10 & 50.69 & 84.76 & 75.59 & 74.61\\

    AAF & 97.86 & 80.40 & 89.44 & 38.38 & 46.33 & 61.19 & \textbf{59.75} & 73.55 & 90.63 & 42.51 & 88.48 & \textbf{81.27} & 80.08 & 93.18 & 89.47 & 93.73 & 60.74 & 86.40 & 75.84 & \textbf{75.22}\\
    \Xhline{1pt}
    \end{tabular}}
    \vspace{0.5pt}
    \caption{Per-class instance-wise IOU results on Cityscapes validation set.}
    \label{tab:cityscapes_inst}
\end{table*}

\subsection{Instance-level Evaluation}
\label{sec:inst_eval}

We measure the instance-wise mIoU on VOC and Cityscapes validation set as summarized in Table~\ref{tab:voc_inst} and Table~\ref{tab:cityscapes_inst}, respectively In instance-wise mIoU, our AAF is higher than base architecture by $3.94\%$ on VOC and $2.94\%$ on Cityscapes. The improvements on fine-structured categories are more prominent. For example, the ``bottle'' is improved by $12.89\%$ on VOC, ``pole'' and ``tlight'' is improved by $9.51\%$ and $9.04\%$ on Cityscapes.

\subsection{Boundary-level Evaluation}
\label{sec:boundary_eval}

\begin{table*}[b!]
  \centering
  \resizebox{\textwidth}{!}{%
  \begin{tabular}{l|c c c c c c c c c c c c c c c c c c c c|c}
    \Xhline{1pt}
    Method & aero & bike & bird & boat & bottle & bus & car & cat & chair & cow & table & dog & horse & mbike & person & plant & sheep & sofa & train & tv & mean \\
    \hline \hline
    PSPNet & .694 & .420 & .658 & .417 & .624 & .626 & .562 & .667 & .297 & .587 & .279 & .667 & .608 & .513 & .554 & .235 & .547 & .413 & .551 & .512 & .527 \\
    
    Affinity & .745 & \textbf{.573} & .708 & .524 & .693 & .678 & .627 & .690 & \textbf{.455} & .620 & .383 & .732 & .655 & .602 & \textbf{.648} & \textbf{.370} & .583 & .546 & .609 & \textbf{.635} & \textbf{.610} \\
    
    AAF & .746 & .559 & .704 & .524 & .684 & .675 & .622 & .701 & .441 & .612 & .391 & .728 & .653 & .595 & .647 & .355 & .580 & .547 & .608 & .628 & .606 \\
    \Xhline{1pt}
    \end{tabular}}
    \vspace{0.5pt}
    \caption{Per-class boundary recall results on Pascal VOC 2012 validation set.}
    \label{tab:voc_boundary}
\end{table*}

\begin{table*}[b!]
  \centering
  \resizebox{\textwidth}{!}{%
  \begin{tabular}{l|c c c c c c c c c c c c c c c c c c c|c}
    \Xhline{1pt}
    Method & road & swalk & build. & wall & fence & pole & tlight & tsign & veg. & terrain & sky & person & rider & car & truck & bus & train & mbike & bike & mean \\
    \hline \hline
    
    PSPNet & .458 & .771 & .584 & .480 & .537 & .587 & .649 & .687 & .650 & .589 & .587 & .733 & .631 & .812 & .577 & .734 & .569 & .550 & .697 & .625 \\
    \hline

    Affinity & .484 & .826 & .686 & .532 & .632 & \textbf{.760} & \textbf{.769} & \textbf{.780} & .754 & .663 & .655 & \textbf{.814} & \textbf{.748} & .852 & .627 & .792 & .589 & .651 & .798 & \textbf{.706}\\

    AAF & .482 & .826 & .685 & .533 & .643 & .756 & .768 & .780 & .753 & .645 & .653 & \textbf{.814} & .746 & .851 & .644 & .789 & .590 & .642 & \textbf{.801} & .705\\
    \Xhline{1pt}
    \end{tabular}}
    \vspace{0.5pt}
    \caption{Per-class boundary recall results on Cityscapes validation set.}
    \label{tab:cityscapes_boundary}
\end{table*}

Next, we analyze quantitatively the improvements of boundary localization. We include the boundary recall on VOC in Table~\ref{tab:voc_boundary} and Cityscapes in Table~\ref{tab:cityscapes_boundary}. We omit the precision table due to smaller performance difference. The overall boundary recall is improved by $7.9\%$ and $8.0\%$ on VOC and Cityscapes, respectively. It is worth noting that the boundary recall is improved for every category. This result demonstrates that boundaries of all categories can all benefit from affinity fields and AAF. Among all, the improvements on categories with complicated boundaries, such as ``bike'', ``bird'', ``boat'', ``chair'',  ``person'', and ``plant'' are significant on VOC. On Cityscapes, objects with thin structures are improved most, such as ``pole'', ``tlight'', ``tsign'', ``person'', ``rider'', and ``bike''.

\subsection{Adaptive Affinity Field Size Analysis}
\label{sec:aaf_analysis}
We further analyze our proposed AAF methods on: 1) optimal affinity field size for each category, and 2) effective combinations of affinity field sizes.

\begin{figure}[t]
    \centering
    \includegraphics[width=0.99\linewidth]{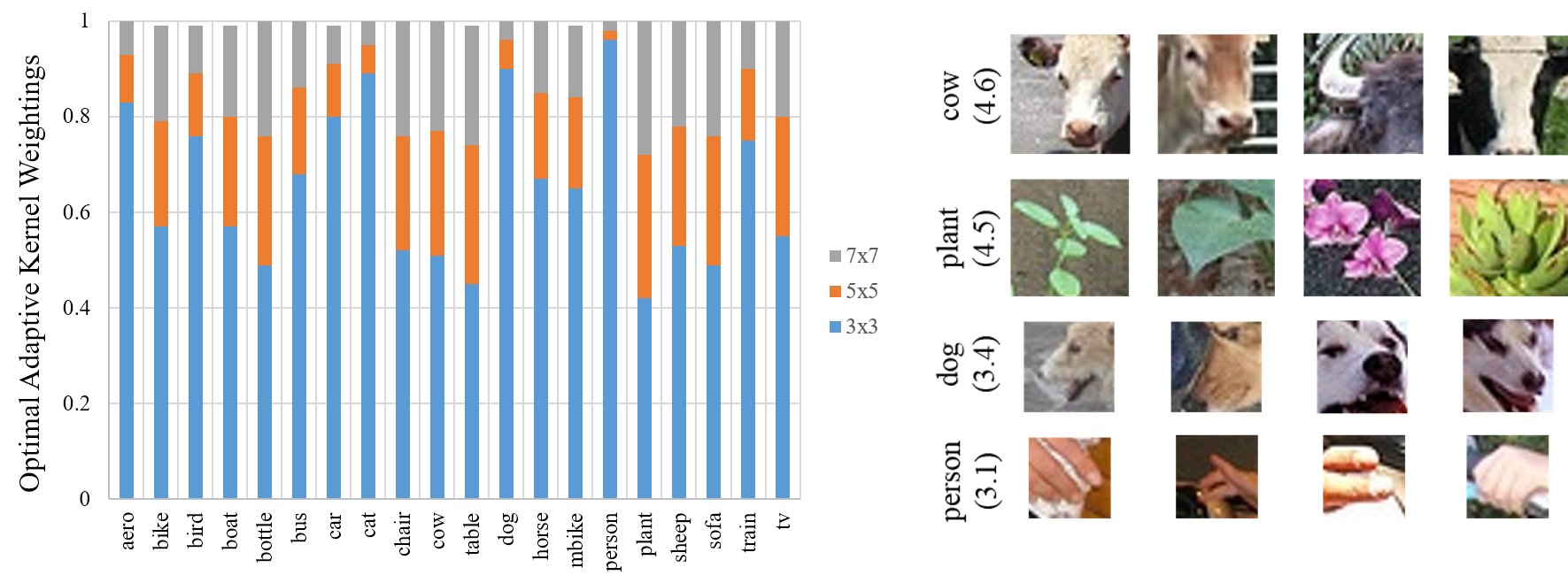}
    \caption{{\bf Left:} The optimal weightings for different kernel sizes of the edge term in AAF for each category on PASCAL VOC 2012 validation set.  {\bf Right:} Visualization of image patches with corresponding effective receptive field sizes, suggesting how kernel sizes capture the shape complexity in critical regions of different categories.}
    \label{fig:aff_size}
\end{figure}

\noindent
\textbf{Optimal Adaptive Affinity Field Size.}
We conduct experiments on VOC with our proposed AAF on three $k \times k$ kernel sizes where $k=3,5,7$. We report the optimal adaptive kernel size on the contour term calculated as $k^{e}_c=\sum_{k} w_{eck} \times k$, and summarized in Fig.~\ref{fig:aff_size}. As shown, ``person'' and ``dog'' benefit from smaller kernel size ($3.1$ and $3.4$), while ``cow'' and ``plant''from larger kernel size ($4.6$ and $4.5$). We display some image patches with the corresponding effective receptive field size.

\noindent
\textbf{Combinations of Affinity Field Sizes.}
We explore the effectiveness of different selections of $k \times k$ kernels, where $k \in \{3,5,7\}$, for AAF. Summarized in Table~\ref{tab:aaf_ablation}, we observe that combinations of $3 \times 3$ and $5 \times 5$ kernels have the optimal performance. 

\begin{table*}[t]
  \centering
  \resizebox{\textwidth}{!}{%
  \begin{tabular}{c|c|c|c c c c c c c c c c c c c c c c c c c c|c}
    \Xhline{1pt}
    $k=3$ & $k=5$ & $k=7$ & aero & bike & bird & boat & bottle & bus & car & cat & chair & cow & table & dog & horse & mbike & person & plant & sheep & sofa & train & tv & mIoU \\
    \hline\hline
    
    $\checkmark$ & $\times$ & $\times$ & 89.02 & 68.86 & 90.05 & 73.52 & 77.87 & 94.04 & 86.94 & 91.04 & 40.85 & 85.82 & 54.08 & 84.31 & 89.12 & 84.91 & 86.72 & 67.52 & 85.56 & 52.55 & 87.60 & 73.78 & 79.00 \\

    $\checkmark$ & $\checkmark$ & $\times$ & 90.19 & 68.48 & 89.87 & 76.91 & 77.56 & 93.84 & 89.08 & 91.45 & 40.67 & 85.82 & 57.23 & 85.33 & 89.77 & 85.97 & 86.93 & 65.68 & 85.12 & 52.22 & 87.25 & 74.07 & 79.45 \\

    $\checkmark$ & $\checkmark$ & $\checkmark$ & 89.45 & 68.46 & 90.44 & 75.82 & 77.03 & 94.09 & 88.01 & 91.42 & 38.67 & 85.98 & 56.16 & 84.32 & 89.22 & 84.98 & 87.09 & 67.35 & 87.15 & 55.20 & 88.22 & 73.30 & 79.40 \\
    \Xhline{1pt} 
    \end{tabular}}
    \vspace{0.5pt}
    \caption{Per-category IOU results of AAF with different combinations of kernel sizes $k$ on VOC 2012 validation set. `$\checkmark$' denotes the inclusion of respective kernel size as opposed to `$\times$'.}
    \label{tab:aaf_ablation}
\end{table*}

\subsection{Generalizability}
\begin{figure*}[b]
    \centering
    \includegraphics[width=\linewidth]{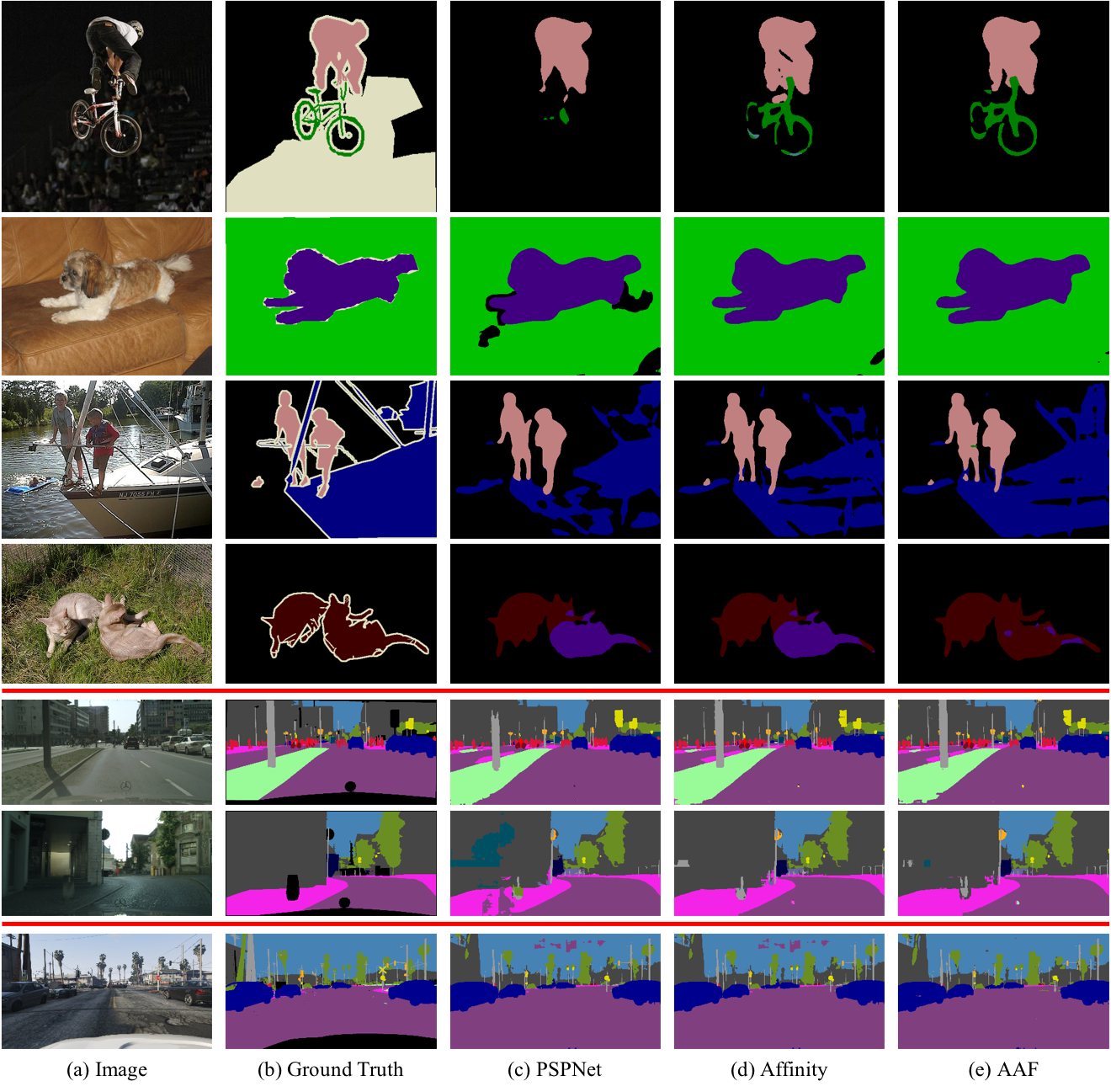}
    \caption{Visual quality comparisons on the VOC 2012~\cite{everingham2010pascal} validation set (the first four rows), Cityscapes~\cite{Cordts2016Cityscapes} validation set (the middle two rows) and GTA5~\cite{Richter_2016_ECCV} part 1 (the bottom row): (a) image, (b) ground truth, (c) PSPNet~\cite{zhao2016pyramid}, (d) affinity fields, and (e) adaptive affinity fields (AAF).}
    \label{fig:results}
\end{figure*}

We further investigate the robustness of our proposed methods on different domains. We train the networks on the Cityscapes dataset~\cite{Cordts2016Cityscapes} and test them on another dataset, Grand Theft Auto V (GTA5)~\cite{Richter_2016_ECCV} as shown in Fig.~\ref{fig:results}. The GTA5 dataset is generated from the photo-realistic computer game--\textit{Grand Theft Auto V}~\cite{Richter_2016_ECCV}, which consists of 24,966 images with densely labelled segmentation maps compatible with Cityscapes. We test on GTA5 Part 1 (2,500 images). We summarize the performance in Table~\ref{tab:gtav}. It is shown that without fine-tuning, our proposed AAF outperforms the PSPNet~\cite{zhao2016pyramid} baseline model by $9.5\%$ in mean pixel accuracy and $1.46\%$ in mIoU, which demonstrates the robustness of our proposed methods against appearance variations. 


\begin{table*}[t]
  \centering
  \resizebox{\textwidth}{!}{%
  \begin{tabular}{l|c c c c c c c c c c c c c c c c c c c |c|c}
    \Xhline{1pt}
    Method & road & swalk & build. & wall & fence & pole & tlight & tsign & veg. & terrain & sky & person & rider & car & truck & bus & train & mbike & bike & mIoU & pix. acc \\
    \hline \hline
    PSPNet & 61.79 & 34.26 & 37.30 & 13.31 & 18.52 & 26.51 & 31.64 & 17.51 & 55.00 & 8.57 & 82.47 & 42.73 & 49.78 & 69.25 & 34.31 & 18.21 & 25.00 & 33.14 & 6.86 & 35.06 & 68.78 \\
    
    Affinity & 75.26 & 30.34 & 44.10 & 12.91 & 20.19 & 29.78 & 31.50 & 23.98 & 64.25 & 11.83 & 74.32 & 48.28 & 49.12 & 67.39 & 25.76 & 23.82 & 20.29 & 41.48 & 5.63 & \textbf{36.86} & 75.13 \\
    
    AAF & 83.07 & 27.82 & 51.16 & 10.41 & 18.76 & 28.58 & 31.74 & 24.98 & 61.38 & 12.25 & 70.65 & 50.53 & 48.06 & 53.35 & 26.80 & 20.97 & 24.50 & 39.56 & 9.37 & 36.52 & \textbf{78.28} \\
    
    \Xhline{1pt}
    \end{tabular}}
    \vspace{0.5pt}
    \caption{Per-class results on GTA5 Part 1.}
    \label{tab:gtav}
\end{table*}

\section{Summary}
\label{sec:conclusion}

We propose adaptive affinity fields (AAF) for semantic segmentation, which incorporate geometric regularities into segmentation models, and learn local relations with adaptive ranges through adversarial training. Compared to other alternatives, our AAF model is 1) effective (encoding rich structural relations), 2) efficient (introducing no inference overhead), and 3) robust (not sensitive to domain changes). 
Our approach achieves competitive performance on standard benchmarks and also generalizes well on unseen data.
It provides a novel perspective towards structure modeling in deep learning.

\clearpage

\bibliographystyle{splncs04}
\bibliography{aseg}



\end{document}